\pdfoutput=1

\documentclass[11pt]{article}

\usepackage[final]{acl}

\usepackage{times}
\usepackage{latexsym}

\usepackage[T1]{fontenc}

\usepackage[utf8]{inputenc}
\usepackage{todonotes}
\usepackage{microtype}

\usepackage{inconsolata}
\usepackage{graphicx}
\usepackage{bm}
\usepackage{svg}
\usepackage{amsmath}
\usepackage{amsfonts}
\usepackage{amssymb}
\usepackage{graphicx}
\usepackage{multirow}
\usepackage{times}
\usepackage{tabularx}
\usepackage{fancyhdr,graphicx,amsmath,amssymb}
\usepackage{wasysym}

\usepackage{algpseudocode}
\usepackage[ruled,linesnumbered,resetcount]{algorithm2e}
\include{pythonlisting}
\usepackage{multirow}
\usepackage{color}
\usepackage{soul}
\usepackage{graphicx}
\usepackage{tablefootnote}

\usepackage{arydshln}
\usepackage{todonotes}

\usepackage{times}
\usepackage{microtype}
\usepackage{hyperref}
\usepackage{url}
\usepackage{booktabs}

\usepackage{lineno}

\definecolor{darkblue}{rgb}{0, 0, 0.5}
\hypersetup{colorlinks=true, citecolor=darkblue, linkcolor=darkblue, urlcolor=darkblue}

\usepackage{latexsym}

\usepackage[T1]{fontenc}

\usepackage[utf8]{inputenc}
\usepackage{todonotes}
\usepackage{microtype}

\usepackage{inconsolata}
\usepackage{graphicx}
\usepackage{bm}
\usepackage{svg}
\usepackage{amsmath}
\usepackage{amsfonts}
\usepackage{amssymb}
\usepackage{graphicx}
\usepackage{multirow}
\usepackage{tabularx}
\usepackage{fancyhdr,graphicx,amsmath,amssymb}
\usepackage{wasysym}
\usepackage{algpseudocode}
\usepackage[ruled,linesnumbered,resetcount]{algorithm2e}
\include{pythonlisting}
\usepackage{multirow}
\usepackage{color}
\usepackage{soul}
\usepackage{graphicx}
\usepackage{tablefootnote}
\usepackage{arydshln}
\usepackage{wrapfig}
\usepackage{listings}
\usepackage{xcolor}

\title{BamiBERT: A New BERT-based Language Model for Vietnamese}

\author{Dat Quoc Nguyen$^1$, Thinh Pham$^2$, Chi Tran$^1$, Linh The Nguyen$^1$\\
$^1$Qualcomm AI Research\thanks{Qualcomm Vietnam Company Limited. Qualcomm AI Research is an initiative of Qualcomm Technologies, Inc. This work was completed while all authors were at Movian AI, Vietnam. All datasets and models were downloaded, trained, and evaluated using Movian AI’s resources.}\\
$^2$Virginia Tech\\
\texttt{\{datnq, chitran, linhnt\}@qti.qualcomm.com, thinhphp@vt.edu}}

\begin{document}

\maketitle
\begin{abstract}
In this paper, we introduce BamiBERT, a new BERT-based pre-trained language model for Vietnamese that addresses key limitations of PhoBERT---the current de facto Vietnamese text encoder. Trained from scratch on a 129GB corpus of general-domain Vietnamese text for 20 epochs, BamiBERT supports an extended context length of up to 2048 tokens and operates directly on raw input, eliminating the need for external word segmentation. Across 8 Vietnamese benchmarks, it achieves the best score on 11 of 15 metrics and the second-best on 3 others, setting a new state of the art among "base"-sized Vietnamese encoders and demonstrating strong cross-domain generalization.
\end{abstract}

\section{Introduction}

In today's LLM-driven era, BERT-based models \citep{devlin-etal-2019-bert, liu2019robertarobustlyoptimizedbert} remain essential for tasks that demand high precision and low latency, such as span labeling, classification, and information retrieval. Their lightweight nature makes them particularly well-suited for resource-constrained applications. Rather than competing with large language models (LLMs), BERT-based models often serve as core components in hybrid systems, delivering strong performance at a fraction of the computational cost while effectively complementing LLMs \cite{10.1145/3637528.3671470}. As a result, the BERT family continues to evolve, with recent additions such as ModernBERT \cite{modernBERT} and NeoBERT \cite{NeoBERT}.

While English benefits from a rich ecosystem of pre-trained BERT-based models, the development of Vietnamese counterparts remains comparatively limited. The multilingual model XLM-RoBERTa \cite{xlmr} achieves competitive performance on a wide range of Vietnamese NLP tasks by leveraging 138GB of CC100 Vietnamese text in its multilingual pre-training corpus. PhoBERT \cite{phobert} was the first large-scale monolingual BERT model pre-trained from scratch specifically for Vietnamese, using 20GB of text. Subsequent general-domain monolingual models include viBERT and vELECTRA \cite{bui-etal-2020-improving}, pre-trained on 60GB of Vietnamese text, and ViDeBERTa \cite{tran-etal-2023-videberta}, which uses the same 138GB CC100 Vietnamese corpus as XLM-RoBERTa. More recently, CafeBERT \cite{do-etal-2024-vlue} was continually pre-trained from the XLM-RoBERTa ``large'' model on 18GB of Vietnamese text collected prior to 2021. Several domain-specific monolingual models have also emerged: VnLawBERT \cite{VNLawBERT} for legal text, ViHealthBERT \citep{vihealthbert} and ViPubmedDeBERTa \cite{vipubmeddeberta} for health and biomedical text, and ViSoBERT \cite{nguyen-etal-2023-visobert} for social media text. 

Among these monolingual models, PhoBERT has become the default choice for many Vietnamese NLP tasks thanks to its strong and consistent performance. Since its release, it has gained widespread adoption, with over 200K monthly downloads on HuggingFace and active use across the NLP community, whereas all other Vietnamese monolingual models receive fewer than 5K monthly downloads each. Despite its popularity, PhoBERT has several limitations: it supports only a short maximum context length of 256 subword tokens, and it requires input text to be word-segmented by an external tool prior to processing. These limitations motivate the development of a new Vietnamese BERT-based model that supports a longer context and operates directly on raw text.

In this paper, we introduce BamiBERT---a new pre-trained language model for Vietnamese---trained from scratch on a large corpus of 129GB of uncompressed text for 20 epochs, with an extended maximum context length of 2048 tokens. Unlike PhoBERT, which requires Vietnamese text to be pre-segmented by an external word segmenter, BamiBERT operates directly on raw input text, making it more flexible and easier to integrate into a wider range of downstream applications. Experimental results on 8 Vietnamese benchmark datasets show that BamiBERT delivers state-of-the-art or near-state-of-the-art performance (ranking \#1 on 11/15 metrics, \#2 on 3/15 metrics, and \#3 on the remaining one), demonstrating strong cross-domain generalization. We release BamiBERT at: \url{https://huggingface.co/Qualcomm-AI-Research/BamiBERT}.

\section{Pre-trained language model  BamiBERT}\label{ssec:BamiBERT}

This section presents how we pre-train our new BERT-based language model from scratch.

\paragraph{Architecture:} We pre-train a text encoder, named "BamiBERT",\footnote{"Bami" denotes "bánh mì" which is a popular type of sandwich in Vietnam.} 
 from scratch, employing the BERT's "base" architecture with 12 Transformer block layers \citep{devlin-etal-2019-bert}. To pre-train BamiBERT, we use the masked language modeling objective \citep{devlin-etal-2019-bert} and the RoBERTa pre-training approach \citep{liu2019robertarobustlyoptimizedbert} which optimizes BERT with a dynamic masking strategy and without the next sentence prediction objective. 
For tokenization, we extend the PhoGPT's Vietnamese-specific byte-level BPE tokenizer \citep{phogpt} with an additional "<mask>" token, resulting in a final vocabulary of 20481 token types. We set a maximum sequence length of 2048.

\paragraph{Pre-training dataset:} We use a clean, 129 GB dataset of uncompressed, general-domain text. 

\paragraph{Optimization:} The model is optimized using Adam \citep{KingmaB14}. We use a batch size of 1024 sequence blocks distributed across 8 A100 GPUs (each with 40GB of memory) and a peak learning rate of 0.00015. The pre-training process runs for 20 epochs, with the initial 2 epochs dedicated to warming up the learning rate.

\section{Experiments}

\subsection{Setup}

\begin{table}[!t]
    \centering\resizebox{7.5cm}{!}{
    \begin{tabular}{l|l|l|l}
    \hline
       \textbf{Dataset}   &  \textbf{Train.} & \textbf{Valid.}  & \textbf{Test} \\
       \hline
       ViNLI  & 24,376 & 3,009 & 2,991  \\
       \hline
         PhoNER\_COVID19 &  5,027 & 2,000 & 3,000   \\
        \hline
        UIT-VSFC (Sentiment) & 11,426 &	1,583 & 3,166 \\
        \hline
        UIT-VSFC (Topic) & 11,426 &	1,583 & 3,166 \\
        \hline
        ViSpamReviews  & 14,306 & 1,590 & 3,974 \\
        \hline
        UIT-ViSFD & 7,786 & 1,112 & 2,224\\
        \hline
        UIT-ABSA (Hotel) & 7,180 & 795 & 2,030\\
        \hline
        UIT-ABSA (Restaurant) & 7,028 & 771 & 1,938 \\
        \hline
    \end{tabular}
    }
    \caption{Statistics of 8 experimental datasets.}
    \label{tab:datasets}
\end{table}

\begin{table*}[!t]
\centering
\resizebox{16cm}{!}{
\setlength{\tabcolsep}{0.3em}
\def\arraystretch{1.1}
\begin{tabular}{l | l | c | c | c | c | c }
\hline
Dataset & Metric & BamiBERT & ViDeBERTa & ViSoBERT & XLM-RoBERTa & PhoBERT \\
\hline 
\multirow{2}{*}{ViNLI (4-label)} & Accuracy & \textbf{81.01} & 61.08 & 67.70 & 76.83$^\dagger$ & \underline{78.00}$^\dagger$ \\
\cdashline{2-7}
 & F\textsubscript{1} & \textbf{81.15} & 60.71 & 67.82 & 77.01$^\dagger$ & \underline{78.05}$^\dagger$ \\
\hline
PhoNER\_COVID19 & F\textsubscript{1} & \textbf{94.90} & \underline{94.50}$^\dagger$ & 92.90 & 92.50$^\dagger$ & 94.20$^\dagger$ \\
\hline
\multirow{2}{*}{UIT-VSFC (Sentiment)} & Accuracy & \underline{93.86} & 87.86 & 93.15 & 93.56 & \textbf{94.10} \\
\cdashline{2-7}
 & F\textsubscript{1} & \textbf{83.41} & 70.74 & 81.49 & 82.20 & \underline{83.27} \\
\hline
\multirow{2}{*}{UIT-VSFC  (Topic)} & Accuracy & \textbf{89.34} & 83.94 & 88.80 & 89.18 & \underline{89.24} \\
\cdashline{2-7}
 & F\textsubscript{1}  & \textbf{79.90} & 66.49 & \underline{79.86} & 79.56 & 79.80 \\
\hline
\multirow{2}{*}{ViSpamReviews}  & Accuracy & \underline{90.76} & 86.21 &\textbf{ 90.99}$^\dagger$ & 90.16$^\dagger$ & 89.83$^\dagger$ \\
\cdashline{2-7}
 & F\textsubscript{1} & \underline{78.20} & 67.04 & \textbf{79.06}$^\dagger$ & 76.55$^\dagger$ & 76.18$^\dagger$ \\
\hline
\multirow{2}{*}{UIT-ViSFD}  & F\textsubscript{1} (Detection) & \textbf{89.14} & 75.53 & \underline{88.63} & 82.73 & 86.03 \\
\cdashline{2-7}
 & F\textsubscript{1} (Classification) & \textbf{84.24} & 64.20 & \underline{83.55} & 65.81 & 78.76 \\
\hline
\multirow{2}{*}{UIT-ABSA (Hotel)}  & F\textsubscript{1} (Detection) & \textbf{79.99} & 72.05 & \underline{79.41} & 77.70$^\dagger$ & 79.16$^\dagger$ \\
\cdashline{2-7}
 & F\textsubscript{1} (Classification) & 72.65 & 62.97 & \textbf{74.24} & 71.23$^\dagger$ & \underline{73.73}$^\dagger$ \\
\hline
\multirow{2}{*}{UIT-ABSA (Restaurant)} & F\textsubscript{1} (Detection) & \textbf{88.01} & 73.56 & \underline{86.86} & 82.18$^\dagger$ & 86.53$^\dagger$ \\
\cdashline{2-7}
 & F\textsubscript{1} (Classification) & \textbf{74.89} & 63.78 & \underline{73.87} & 71.58$^\dagger$ & 73.52$^\dagger$ \\
\hline 
\end{tabular}
}
\caption{Results of pre-trained "base"-architecture models. $^\dagger$ denotes results extracted from previous works.}
\label{tab:results}
\end{table*}

We conduct experiments to compare our model BamiBERT  with the previous strong and public pre-trained "base"-architecture ones for Vietnamese, including: Vietnamese-specific models ViDeBERTa-base \citep{tran-etal-2023-videberta}, ViSoBERT \citep{nguyen-etal-2023-visobert} and PhoBERT-base \citep{phobert} as well as the multilingual XLM-RoBERTa-base \citep{xlmr}.\footnote{ViDeBERTa, ViSoBERT and XLM-RoBERTa were trained using a maximum sequence length of 512 tokens.} Here, BamiBERT, ViSoBERT and XLM-RoBERTa take raw texts as input, while ViDeBERTa and PhoBERT are Vietnamese word-level models. That is, a Vietnamese word segmentation tool must be applied to produce word-segmented texts before feeding them to the word-level ViDeBERTa and PhoBERT. For ViDeBERTa and PhoBERT experiments, we utilize the RDRSegmenter component \citep{NguyenNVDJ2018} from the VnCoreNLP toolkit \citep{vu-etal-2018-vncorenlp} for Vietnamese word segmentation. 

We employ the following experimental benchmark datasets: ViNLI---a Vietnamese dataset for open-domain natural language inference \citep{huynh-etal-2022-vinli}, PhoNER\_COVID19---a dataset for recognizing COVID-19 related named entities in Vietnamese \citep{PhoNER_COVID19}; UIT-VSFC (Sentiment) and UIT-VSFC (Topic)---Vietnamese students’ feedback benchmarks for sentiment-based and topic-based classifications \citep{8573337}; ViSpamReviews---a dataset for spam review detection on Vietnamese e-commerce websites \citep{vispamreviews}; UIT-ViSFD---a Vietnamese aspect-based sentiment analysis dataset of feedbacks and comments for smartphone e-commerce \citep{visfd}; and UIT-ABSA (Hotel) and UIT-ABSA (Restaurant)---Vietnamese aspect-based sentiment analysis datasets for hotel and restaurant domains \citep{absa}. 
ViNLI and PhoNER\_COVID19 are based on general-domain texts, whereas the remaining benchmarks are derived from social media and forum discussions. 
See Table \ref{tab:datasets} for the statistics of these datasets. 

For all experimental models, we employ \texttt{transformers} \citep{wolf-etal-2020-transformers} to fine-tune them using the AdamW optimizer \citep{loshchilov2018decoupled} and set the batch size to 32. We also perform a grid search on the validation set to select the initial learning rate for AdamW from \{1e-5, 2e-5, 5e-5\}. We train for 30 epochs on the training set, compute F\textsubscript{1} on the validation set after each training epoch, and select the model checkpoint with the best F\textsubscript{1} to report final metric scores on the test set.

\subsection{Main Results}
\label{sec:main_results}

Table~\ref{tab:results} reports the performance of BamiBERT and four
baselines---ViDeBERTa, ViSoBERT, XLM-RoBERTa, and PhoBERT---across eight
Vietnamese benchmarks.  BamiBERT achieves the best performance on 11 of the 15 evaluation metrics, ranks second on three metrics, and ranks third on the remaining metric, establishing a new state of the
art for "base"-sized Vietnamese BERT-based language models. 

\paragraph{ViNLI}

BamiBERT achieves the best performance on both metrics (81.01 Accuracy and 81.15 F\textsubscript{1}), yielding substantial absolute gains of +3.01 Accuracy and +3.10 F\textsubscript{1} over the second-ranked PhoBERT (78.00/78.05). XLM-RoBERTa follows in third place (76.83/77.01), trailing PhoBERT by roughly one point on both metrics. ViSoBERT (67.70/67.82) and ViDeBERTa (61.08/60.71) lag considerably behind, with gaps of 13--20 points relative to BamiBERT.

\paragraph{PhoNER\_COVID19}

BamiBERT obtains the highest F\textsubscript{1} score (94.90), outperforming ViDeBERTa by 0.40 points and PhoBERT by 0.70 points. Its advantage widens against the remaining baselines, reaching +2.0 F\textsubscript{1} over ViSoBERT (92.90) and +2.4 F\textsubscript{1} over XLM-RoBERTa (92.50).

\paragraph{UIT--VSFC (Sentiment)}
With 93.86 Accuracy and 83.41 F\textsubscript{1}, BamiBERT ranks among the top-performing models. It is essentially on par with PhoBERT, trailing by 0.24 Accuracy but leading by 0.14 F\textsubscript{1}. BamiBERT maintains consistent margins over XLM-RoBERTa (+0.30 Accuracy, +1.21 F\textsubscript{1}) and ViSoBERT (+0.71 Accuracy, +1.92 F\textsubscript{1}), while ViDeBERTa underperforms substantially on both metrics.

\paragraph{UIT--VSFC (Topic)} 
BamiBERT again leads on both metrics (89.34 Accuracy and 79.90 F\textsubscript{1}), outperforming PhoBERT by +0.10 on both, XLM-RoBERTa by +0.16/+0.34, and ViSoBERT by +0.54/+0.04. Although the top four models are tightly clustered (within 0.54 Accuracy and 0.34 F\textsubscript{1}), BamiBERT remains the most consistent. ViDeBERTa, in contrast, trails markedly (83.94/66.49).

\paragraph{ViSpamReviews}

BamiBERT ranks second overall (90.76 Accuracy and 78.20 F\textsubscript{1}), narrowly behind ViSoBERT (90.99/79.06) by 0.23 Accuracy and 0.86 F\textsubscript{1}. Nevertheless, it clearly outperforms XLM-RoBERTa (+0.60 Accuracy, +1.65 F\textsubscript{1}) and PhoBERT (+0.93 Accuracy, +2.02 F\textsubscript{1}), while ViDeBERTa lags well behind (86.21/67.04). BamiBERT remains highly competitive at the top tier and continues to surpass other established baselines by clear margins.

\paragraph{UIT--ViSFD}

BamiBERT delivers the strongest performance on both subtasks. For aspect detection, it attains 89.14 F\textsubscript{1}, ahead of ViSoBERT (88.63; $-$0.51) and PhoBERT (86.03; $-$3.11), with XLM-RoBERTa (82.73) and ViDeBERTa (75.53) trailing further. For aspect-based sentiment classification, BamiBERT achieves 84.24 F\textsubscript{1}, again surpassing ViSoBERT (83.55; $-$0.69) and PhoBERT (78.76; $-$5.48). The consistent gains over PhoBERT (+3.11 and +5.48 points) underscore BamiBERT's robustness across both subtasks.

\paragraph{UIT--ABSA (Hotel)}
On aspect detection, BamiBERT obtains the highest F\textsubscript{1} (79.99), outperforming ViSoBERT (79.41; $-$0.58) and PhoBERT (79.16; $-$0.83), while XLM-RoBERTa (77.70) and ViDeBERTa (72.05) remain less competitive. On aspect-based sentiment classification, however, ViSoBERT takes the lead (74.24 F\textsubscript{1}), followed by PhoBERT (73.73; $-$0.51) and BamiBERT (72.65; $-$1.59); XLM-RoBERTa (71.23) and ViDeBERTa (62.97) trail by a wide margin.

\paragraph{UIT--ABSA (Restaurant)}
BamiBERT produces the strongest end-to-end performance, with 88.01 F\textsubscript{1} on aspect detection and 74.89 F\textsubscript{1} on aspect-based sentiment classification. These results exceed those of ViSoBERT (86.86/$-$1.15 and 73.87/$-$1.02) and PhoBERT (86.53/$-$1.48 and 73.52/$-$1.37), while XLM-RoBERTa (82.18/71.58) and ViDeBERTa (73.56/63.78) lag considerably behind.

\section{Discussion}

\textbf{Overall performance}\ \ \  Across 8 Vietnamese benchmarks and
different subtasks (Table~\ref{tab:results}), BamiBERT consistently
delivers SOTA or near-SOTA results. The most substantial improvement
appears on ViNLI, where BamiBERT surpasses the next-best PhoBERT by +3.01
Accuracy and +3.10 F\textsubscript{1}, while outpacing ViSoBERT/ViDeBERTa
by 13--20 points---demonstrating strong sentence-pair semantics and
cue-word sensitivity.

\paragraph{Domain effects}\ \ \  Note that PhoNER\_COVID19 and
ViNLI represent general-domain text, while the remaining benchmarks reflect
social media and forum content. BamiBERT exhibits strong cross-domain
generalization, outperforming or closely matching the social-media-focused
ViSoBERT across both general-domain (e.g., ViNLI, PhoNER\_COVID19) and
social-domain benchmarks. Its consistent top-tier performance across
diverse tasks---NER, span detection, and classification---demonstrates
resilience to domain shift and label granularity. This robustness positions
BamiBERT as a reliable choice for NLP pipelines operating under domain
heterogeneity and distributional uncertainty.

\paragraph{Detection vs.\ classification}\ \ \ In aspect-based
sentiment analysis pipelines, BamiBERT often excels in \textit{detection}
(e.g., Hotel: 79.99 F\textsubscript{1}; Restaurant: 88.01
F\textsubscript{1}), while its \textit{classification} performance is
strongest in the Restaurant domain (74.89 F\textsubscript{1}) but trails
ViSoBERT/PhoBERT in the Hotel domain (72.65 F\textsubscript{1}). This
pattern suggests complementary strengths: BamiBERT appears particularly
effective at span/target localization and boundary-sensitive cues, whereas
domain-specific sentiment nuances in the Hotel domain may benefit more from
domain-adapted pretraining (ViSoBERT). Combined with the new SOTA results
on UIT-ViSFD, this indicates BamiBERT's robust end-to-end capability for
fine-grained social content analysis.

\paragraph{Takeaway}\ \ \  BamiBERT delivers SOTA or near-SOTA
performance across diverse Vietnamese benchmarks, excelling in NLI, topic
classification, and fine-grained sentiment tasks. Its cross-domain
stability and strong results on both detection and classification make it a
robust default for real-world Vietnamese NLP applications.

\section{Conclusion}

In this paper, we have presented {BamiBERT}, a new pre-trained language model for
Vietnamese designed to address key limitations of existing monolingual
encoders. Unlike PhoBERT---the current de facto choice of Vietnamese
text encoder---BamiBERT is trained from scratch on a 129 GB corpus of general-domain text for 20 epochs, supports an extended maximum context length of
2048 tokens, and operates directly on raw text, removing the dependency
on external word segmentation. Experiments on 8 Vietnamese benchmark datasets show that BamiBERT does better than PhoBERT, establishing a new state of the art among ``base''-sized Vietnamese encoders with strong cross-domain generalization.





\bibliography{custom}

\end{document}